\documentclass[conference]{IEEEtran}
\IEEEoverridecommandlockouts
\usepackage{cite}
\usepackage{amsmath,amssymb,amsfonts}
\usepackage{algorithmic}
\usepackage{graphicx}
\usepackage{textcomp}
\usepackage{multirow}
\usepackage{xcolor}
\def\BibTeX{{\rm B\kern-.05em{\sc i\kern-.025em b}\kern-.08em
    T\kern-.1667em\lower.7ex\hbox{E}\kern-.125emX}}
\begin{document}

\title{Stress Classification and Personalization: Getting the most out of the least}

\author{\IEEEauthorblockN{Ramesh Kumar Sah}
\IEEEauthorblockA{\textit{School of Electrical Engineering and Computer Science} \\
\textit{Washington State University}\\
ramesh.sah@wsu.edu}
\and
\IEEEauthorblockN{Hassan Ghasemzadeh}
\IEEEauthorblockA{\textit{School of Electrical Engineering and Computer Science} \\
\textit{Washington State University}\\
hassan.ghasemzadeh@wsu.edu}}

\maketitle

\vspace{-2.5mm}
\begin{abstract}
Stress detection and monitoring is an active area of research with important implications for the personal, professional, and social health of an individual. Current approaches for affective state classification use traditional machine learning algorithms with features computed from multiple sensor modalities. These methods are data-intensive and rely on hand-crafted features which impede the practical applicability of these sensor systems in daily lives. To overcome these shortcomings, we propose a novel Convolutional Neural Network (CNN) based stress detection and classification framework without any feature computation using data from only one sensor modality. Our method is competitive and outperforms current state-of-the-art techniques and achieves a classification accuracy of $92.85\%$ and an $f1$ score of $0.89$. Through our leave-one-subject-out analysis, we also show the importance of personalizing stress models.

\end{abstract}


\vspace{-2.5mm}
\section{Introduction}
\begin{figure}[!tbh]
    \centering
    \includegraphics[width=\linewidth]{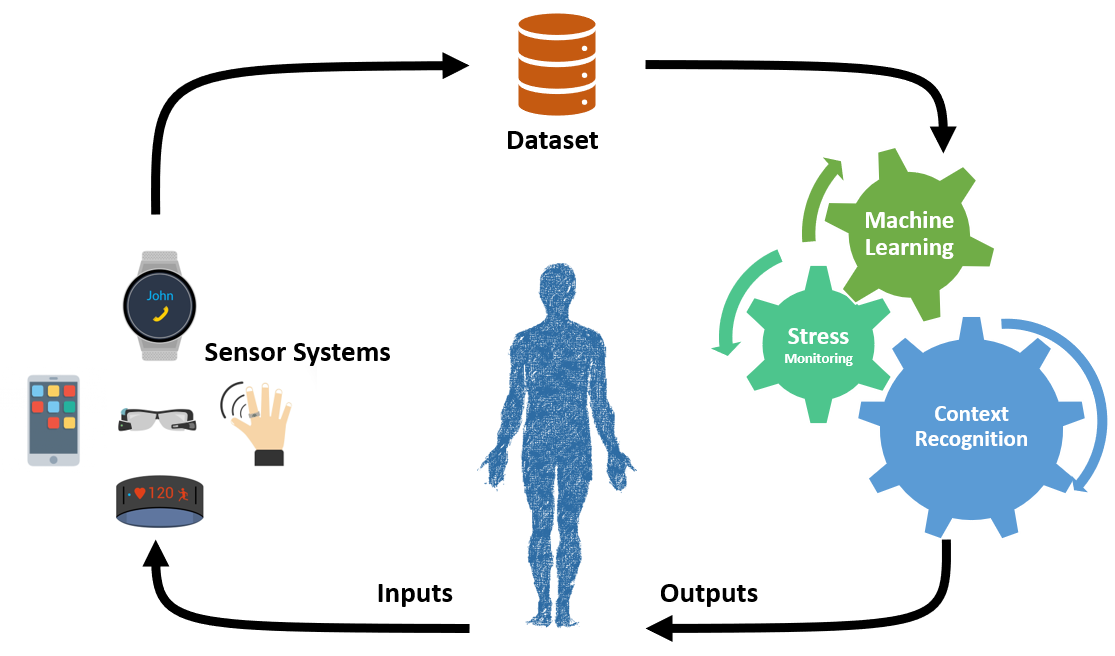}
    \caption{A stress detection and context recognition framework using sensor systems and machine learning algorithms.}
    \label{fig:system}
\end{figure}

Stress describes bodily reactions to perceived physical or psychological threats \cite{b0} and is defined as the transition from a calm state to an excited state triggering a cascade of physiological response \cite{b1}. In the United States of America, around $77\%$ people suffer from headaches and insomnia for reasons related to stress, and there has been a steady increase in the number of people suffering from stress-related issues each year \cite{b2}. %According to the 2017 report by the American Psychological Association, the most common reasons for stress are money, work, violence/crime, and politics \cite{b2}. 
Furthermore, stress plays a critical role in many health problems, such as depression, anxiety, high blood pressure, heart attacks, and stroke \cite{b3}. Stress also influences a person's decision-making capability, attention span, learning, and problem-solving capacity \cite{b4}. Stress detection and monitoring is an active research area with important implications for the personal, professional, and social health of an individual. Stress detection and monitoring can help prevent dangerous stress-related diseases. Towards this end, in this paper we propose, a novel Convolutional Neural Network (CNN) based framework for stress detection and classification, which uses raw Electrodermal Activity (EDA) sensor data without feature computation. Our approach is competitive with other state-of-the-art methods and does not suffers from many limitations inherent in earlier works. Figure \ref{fig:system} shows the general overview of a stress detection system used for real-time interventions to support the health of an individual. In this paper, we implement the stress classification pipeline, and in the future, we aim to use our classification model for strategic real-time interventions.

\label{related_work}
Usually, for stress detection and classification data from multiple sensor modalities such as heart rate variability (HRV), body acceleration (ACC), skin temperature, electrodermal activity (EDA), blood volume pulse (BVP), respiration rate, and electrocardiogram (ECG) are used to compute a large number of statistical and structural features to train machine learning algorithms. In \cite{b6}, the authors computed $67$ features from $7$ sensor modalities to train a stress classification model with the best accuracy of $92.28\%$. Using the same dataset, the authors in \cite{b3} used Deep Neural Networks (DNN) and $40$ statistical features to achieve an accuracy of $95.21\%$. In \cite{b4} the authors used statistical features and representation learned by a deep learning model as features to train the stress classification model with accuracy up to $92\%$ with just EDA data. Motivated by the results from \cite{b6}, the authors in \cite{b7} computed $195$ features in time, frequency, entropy, and wavelet domain from EDA data to train the XGBoost algorithm with the highest accuracy of $92\%$. Feature selection was used to reduce the number of features to $9$ for the best possible classification accuracy. Furthermore, some works have also explored ways not to use electrodermal activity for stress classification since most commercial smartwatches and smart health devices don't have sensors to measure galvanic skin response. In \cite{b8}, authors used data from the built-in smartphone accelerometer sensor to identify activity that corresponds with stress levels and achieved an accuracy of $71\%$. Also, in \cite{b9} data from a commercial smartwatch was used for binary stress classification with accuracy up to $83\%$.

Using data from multiple sensors and computing a large number of features to train machine learning algorithms for stress classification has several disadvantages. Using numerous sensor modalities makes the system design complicated and expensive and hence unfit to be used in everyday lives. Also, sensors need power to operate, and more sensors draw more power, which is a big issue in battery-powered wearable systems. Computing features require domain knowledge, and extensive testing is needed to find the best set of features for optimal classification performance. Furthermore, computing a large number of complex features makes the classification algorithm less efficient in terms of run-time, energy, and memory. Besides, feature selection is needed to select the most meaningful features and adds an extra processing step to an already complex machine learning pipeline. Motivated by these drawbacks of multi-modal feature-based stress classification algorithms, in this paper, we propose a CNN-based stress detection and classification system which takes raw EDA sensor segments as inputs and learns and select the dominant features automatically during the training process. Our primary objective is to implement a stress detection and classification system using only the EDA data. The secondary goal was to explore the personalization of stress models. Perception and effects of stress are subjective in nature. The same external stimuli can have a varying degree of effect on different individuals in terms of stress and emotional arousal. Hence, we also investigate whether stress detection algorithms need personalization or not.

\section{Methodology}
% Our primary objective in this study was to implement a stress detection and classification system using only the Galvanic Skin Response (GSR) data. We wanted to investigate whether raw GSR data can be used to train machine learning algorithms for stress classification without any pre-processing and feature extraction. The secondary objective was to explore the personalization of stress models. Perception and effects of stress is subjective in nature. A same external stimuli can have varying degree of affect on different individuals in terms of stress and emotional arousal. Hence, we also wanted to investigate whether stress detection algorithms need to be personalized for maximum performance or not. We answer these questions using the Wearable Stress and Affect Detection (WESAD) dataset.

\subsection{Dataset}
The Wearable Stress and Affect Detection (WESAD) dataset \cite{b6} is a publicly available dataset with ECG, EDA, BVP, respiration (RESP), skin temperature (TEMP), and motion (Acceleration) (ACC) sensor data obtained from the RespiBan (chest-worn) and Empatica E4 (wrist-worn) devices. The dataset was collected from $15$ subjects (3 female) in a laboratory setting, and each subject experienced three main affect conditions: baseline or normal (neutral reading), stress (exposed to Tier Social Stress Test (TSST)), and amusement (watching funny videos). In our analysis, we only use the EDA data from the Empatica E4 sampled at $4$Hz. Approximately the length of the stressed condition was $10$ minutes, amusement $6.5$ minutes, and baseline situation was $20$ minutes.

\subsection{Segmentation and Normalization}
For each subject, we have approximately $37$ minutes of EDA data. We segment the EDA data for the three affective states into $60$ seconds overlapping segments with $50\%$ overlap between consecutive segments. We settled on the window size of $60$ seconds because of available literature that has also used $60$ seconds window size for the WESAD dataset \cite{b3, b4, b5, b6}. Before segmentation, we normalize the data for each subject using the min-max normalization to spread the data in the range of $[0, 1]$. After segmentation, we obtain $564$ samples for the baseline class, $311$ samples for the stressed class, and $165$ samples for the amusement class. In our analysis, we have not used any method to deal with class imbalance and machine learning models are trained on the imbalance data for the worst case scenario.

\begin{figure}[!tbh]
    \centering
    \includegraphics[width=0.8\linewidth]{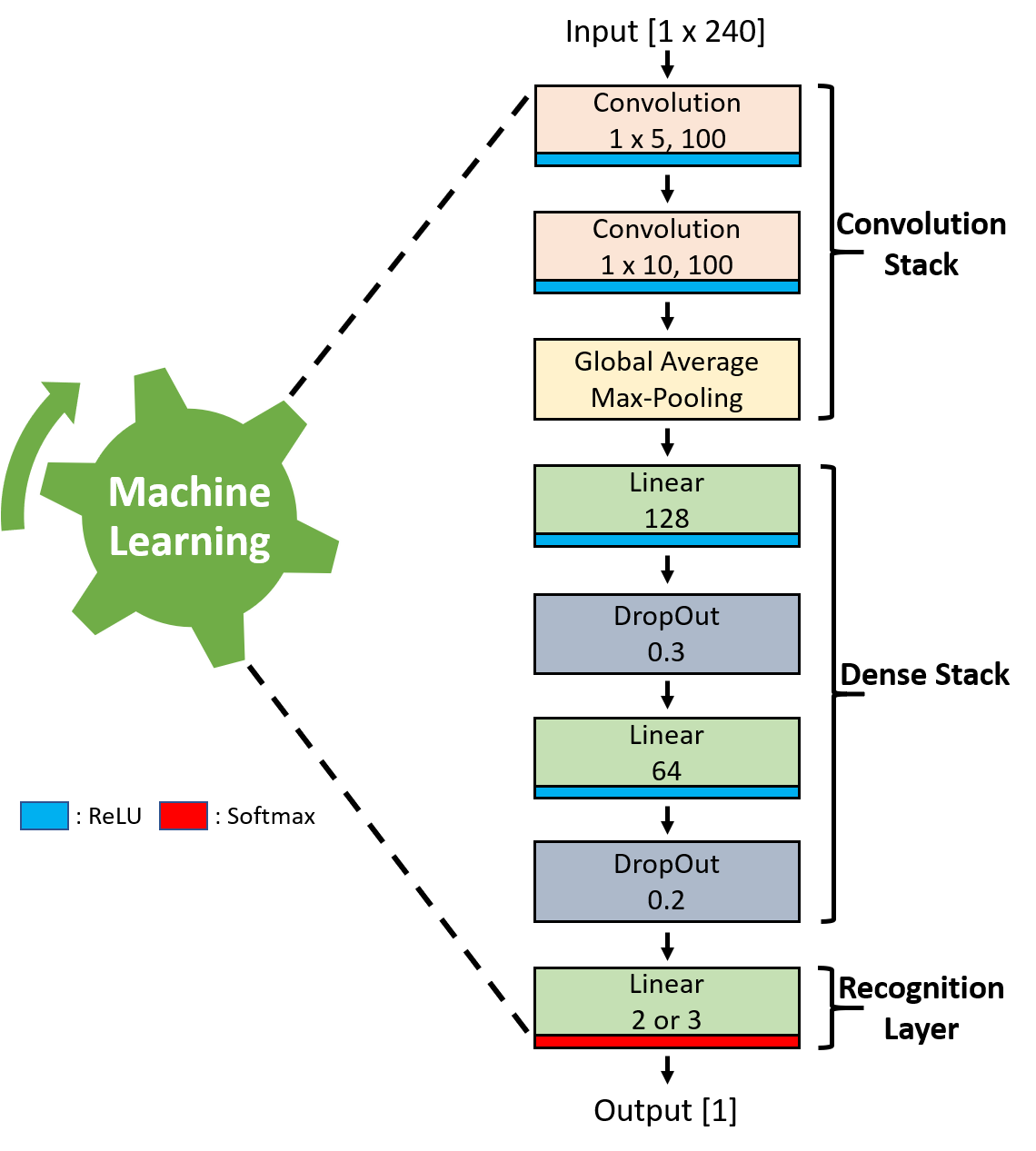}
    \caption{The architecture of the convolutional neural network used for stress classification. Input sensor segment is fed into the convolutional stack to extract features. Extracted features are passed into the dense stack to learn the associations between the input and output classes. The recognition layer predicts stress classes for the input EDA segment.}
    \label{fig:methodology}
\end{figure}
\vspace{-2.5mm}

\subsection{Convolutional Neural Network}
Data-driven learning algorithms learn associations between input and outputs directly from the sensor data without feature computation. These methods learn features and classifier simultaneously from the sensor data. A convolutional neural network (CNN) is a data-driven learning algorithm capable of learning local dependency and scale invariance in the input data without feature computation. In CNN, the convolution operation is used between the input and a weight matrix or filters to assemble complex features by successively learning smaller and simpler features. Consequently, CNN is suitable for our approach towards stress detection and classification, and hence we have used $1D$ CNN as the learning algorithm in our work. We have used a ConvNet architecture composed of two $1D$ convolutional layers with $100$ filters each and kernel size of $5$ and $10$ respectively. This is followed by a global max-pooling layer and two fully connected layers with $128$ and $64$ neurons. We also have drop-out layers after each fully connected layer with drop-out values $0.3$ and $0.2$. The output layer has Softmax activation, and all other layers have ReLU \cite{b10} activation. Figure \ref{fig:methodology} shows the graphical representation of the proposed CNN used in this paper.

\subsection{Hyperparameters and Training}
The hyperparameters in our framework were selected after extensive trial and error. The CNN models were trained for $200$ epochs with a batch size of $32$ and a fixed learning rate of $0.001$. Out of $876$ samples in the dataset, $657$ or $75\%$ was included in the training set, and $219$ or $25\%$ belonged to the test set. For bi-affective state classification, data from the baseline (not-stress) and stressed classes were used to create the training and test sets. For tri-affective state classification data for all three classes: baseline, stressed, and amusement were used to create the training and test sets.

% \subsection{Performance Measure}
% We used the following $4$ performance measures to evaluate the trained CNN models for stress classification in various cases.
% \begin{enumerate}
%     \item Classification Accuracy : The percentage of samples that were correctly classified.
    
%     \item Precision : Precision is the ability of the classifier not to label as positive a sample that is negative. 
    
%     \item Recall : Recall is the ability of the classifier to find all positive sample.
    
%     \item $f1-$score : $f1-$score is the weighted average of precision and recall.
% \end{enumerate}

\section{Results and Observations}
Due to the lack of space, we have omitted training curves of the CNN models and we want to confirm we observed no overfitting during training. 

\subsection{Stress Classification}
First, we present the results for the bi-affective state classification i.e., the binary case of stress Vs. not-stress classification. The trained CNN model achieved the best classification accuracy of $94.8\%$ on the training set and $90.9\%$ on the test set. Table \ref{tab:bi-affective}, shows the value of other performance metrics. 

\vspace{-2.5mm}
\begin{table}[htbp]
    \caption{Results for the bi-affective stress state classification.}
    \begin{center}
    % \resizebox{\linewidth}{!}{%
    \begin{tabular}{|c|c|c|c|c|}
    \hline
    %  & \multicolumn{4}{c|}{\textbf{Bi-affective State Classification}} \\
    % \cline{2-5} 
    Dataset & Accuracy & Precision & Recall & f1-Score \\
    \hline
    Training Set & $94.8\%$ & $0.96$ & $0.88$ & $0.92$  \\
    \hline
    Testing Set & $90.9\%$ & $0.91$ & $0.82$ & $0.87$  \\
    \hline
    \end{tabular}
    % }
    \label{tab:bi-affective}
    \end{center}
\end{table}
\vspace{-2.5mm}

In the second case, we consider the tri-affective state classification, a multi-class classification problem with $3$ classes: stress, not-stress, and amusement. Table \ref{tab:tri-affective} shows the values of performance metrics for this case. Note that the performance of the CNN model has decreased in the tri-affective case compared to the bi-affective case. We suspect this is because the model doesn't have enough training samples to learn the distinction between the three classes.

\vspace{-2.5mm}
\begin{table}[htbp]
    \caption{Results for the tri-affective stress state classification.}
    \begin{center}
    % \resizebox{\linewidth}{!}{%
    \begin{tabular}{|c|c|c|c|c|}
    \hline
    %  & \multicolumn{4}{c|}{\textbf{Tri-affective State Classification}} \\
    % \cline{2-5} 
    Dataset & Accuracy & Precision & Recall & f1-Score \\
    \hline
    Training Set & $85.1\%$ & $0.83$ & $0.79$ & $0.80$  \\
    \hline
    Testing Set & $82\%$ & $0.82$ & $0.72$ & $0.76$  \\
    \hline
    \end{tabular}
    % }
    \label{tab:tri-affective}
    \end{center}
\end{table}
\vspace{-2.5mm}

Furthermore, to account for the variance in performance, we conducted $10-$fold cross-validation for both cases of affective state classification. Table \ref{tab:cv_results} shows the average classification accuracy and f1-score for bi-affective and tri-affective cases. 

\vspace{-2.5mm}
\begin{table}[htbp]
    \caption{Average classification accuracy and f1-score.}
    \begin{center}
    % \resizebox{\linewidth}{!}{%
    \begin{tabular}{|c|c|c|c|}
    \hline
     & Dataset & Accuracy & f1-Score \\
    \hline
    \multirow{ 2}{*}{Bi-affective} & Training Set & $93\%$ & $0.9$  \\
     & Testing Set & $90\%$ & $0.86$ \\
    \hline
    \multirow{ 2}{*}{Tri-affective} & Training Set & $84\%$ & $0.79$  \\
     & Testing Set & $80\%$ & $0.75$ \\
    \hline
    \end{tabular}
    % }
    \label{tab:cv_results}
    \end{center}
\end{table}
\vspace{-2.5mm}

% \begin{figure}[!tbh]
%     \centering
%     \includegraphics[width=\linewidth]{images/bi_cv_results_1.pdf}
%     \caption{Average classification accuracy and f1-score from $10-$fold cross-validation for bi-affective stress classification.}
%     \label{fig:bi_cv_results}
% \end{figure}

% \begin{figure}[!tbh]
%     \centering
%     \includegraphics[width=\linewidth]{images/tri_cv_results_1.pdf}
%     \caption{Average classification accuracy and f1-score from $10-$fold cross-validation for tri-affective stress classification.}
%     \label{fig:tri_cv_results}
% \end{figure}

Finally, we present comparisons of our results with other state-of-the-art works on stress classification with the WESAD dataset in table \ref{tab:comp_SOTA}. WESAD dataset has the following modalities ACC, EDA, TEMP, ECG, BVP, and RESP, and is represented by \textit{All} in the table. All other compared approaches, details in \ref{related_work}, computes statistical or representational features from sensor data to train stress classification models. Our method, does not involves computation intensive feature computation and selection stages and uses the raw sensor data for training. Also, our approach is based on CNNs whereas compared methods are based on neural networks as well as statistical learning algorithms. We found our proposed approach to be competitive with state-of-the-art methods with the added advantage of being data-driven without needing any specialized domain knowledge for feature computation and selection.

\vspace{-2.5mm}
\begin{table}[htbp]
    \caption{Comparisons of our proposed approach with state-of-the-art methods.}
    \begin{center}
    % \resizebox{\linewidth}{!}{%
    \begin{tabular}{|c|c|c|c|c|}
    \hline
     Method & Model Type & Modalities & Accuracy (\%) & f1-Score \\
    \hline
    \cite{b6} & Feature & All & $93$ & $0.9$  \\
    \hline
    \cite{b4} & Feature & EDA & $91.60$ & - \\
    \hline
    \cite{b7} & Feature & EDA & $-$ & $0.92$ \\
    \hline
    \cite{b3} & Feature & All & $95.21$ & $0.94$\\
    \hline
    \textbf{Our's} & \textbf{Data} & \textbf{EDA} & \mbox{\boldmath$92.85$} & \mbox{\boldmath $0.89$} \\
    \hline
    \end{tabular}
    % }
    \label{tab:comp_SOTA}
    \end{center}
\end{table}
\vspace{-2.5mm}

\subsection{Personalization of Stress Models}
\begin{figure}[!tbh]
    \centering
    \includegraphics[width=0.9\linewidth]{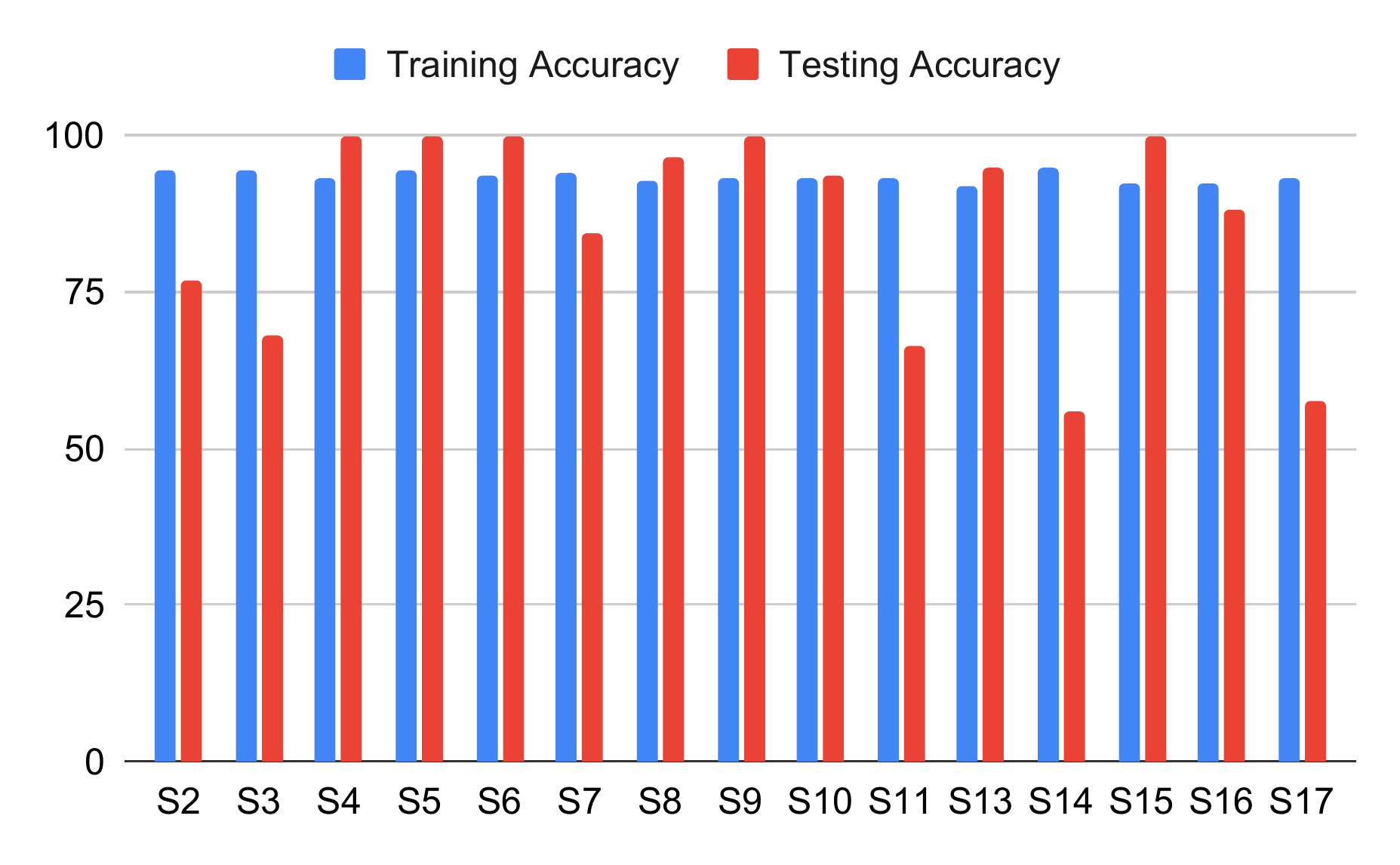}
    \caption{Classification accuracy on the training and test sets for leave-one-subject-out analysis. The x-axis represents the subject whose data was not included in the training set and was used as the test set. }
    \label{fig:loso_acc}
\end{figure}

\begin{figure}[!tbh]
    \centering
    \includegraphics[width=0.9\linewidth]{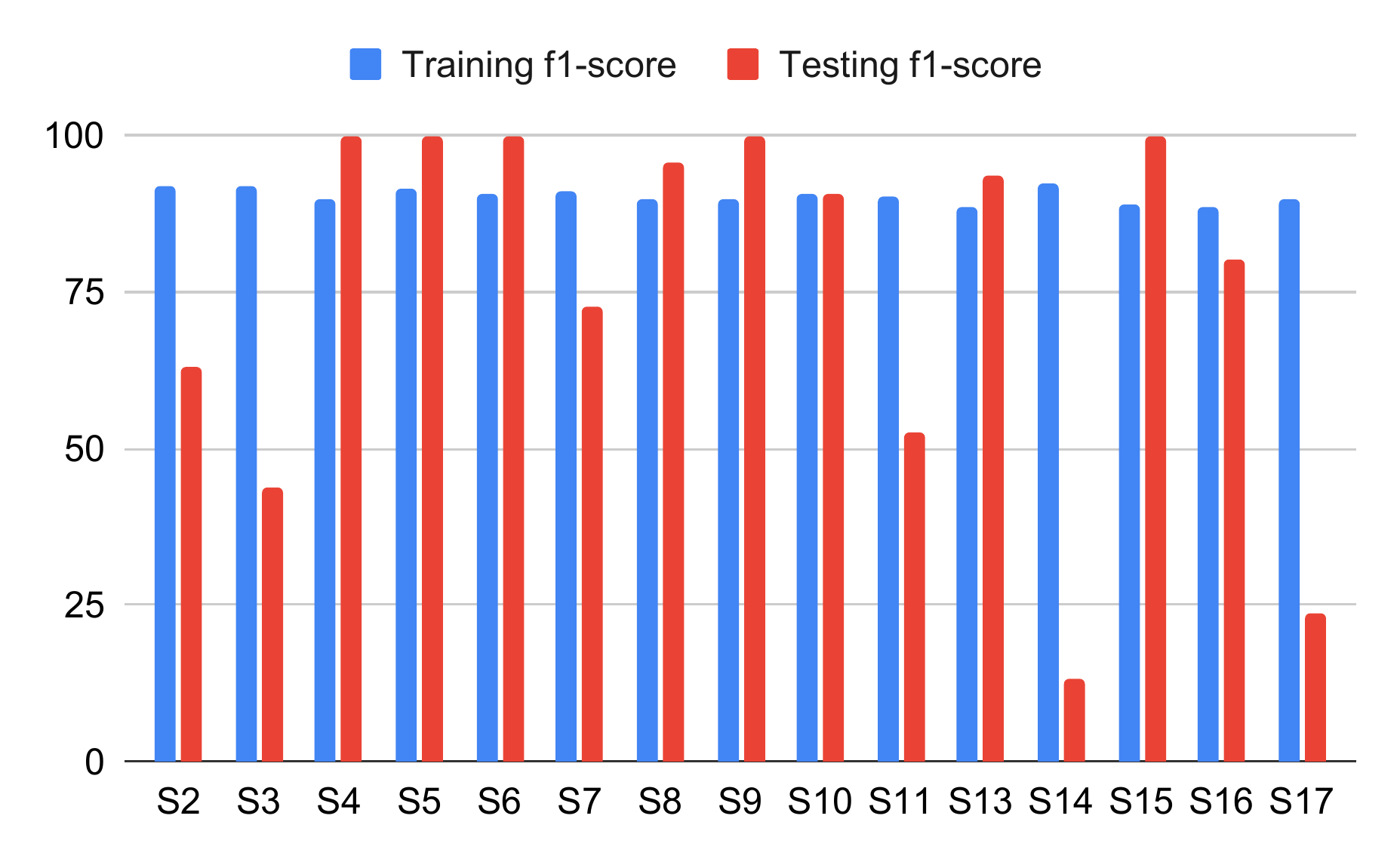}
    \caption{f1-score on the training and test sets for the leave-one-subject-out analysis to investigate the subjective nature of stress. The x-axis represents the subject whose data was not included in the training set and was used as the test set.}
    \label{fig:loso_f1}
\end{figure}

To investigate the subjective nature of stress and determine whether we need personalized models for stress detection and classification, we present the results of leave-one-subject-out (LOSO) analysis on the binary WESAD dataset. In LOSO analysis, data from one subject is removed from the training set and kept as the test set to evaluate the trained machine learning model. The WESAD dataset was collected from $15$ subjects and we present the results of LOSO analysis for each subjects. Figure \ref{fig:loso_acc} shows the classification accuracy and figure \ref{fig:loso_f1} shows the f1-score of the trained models on the test and training sets. The x-axis represents the subject whose data was not included in the training set and was used as the test set. Based on our results, we can confirm that stress is subjective, and the same external stimuli can have varying effects on different individuals. 
%This is reflected in our results, in which we found the trained models to have lower, similar, and higher performance on the test set. 
On the test data for left out subjects \textit{S2, S3, S7, S11, S14,} and \textit{S17} the trained models performed poorly, but for subjects \textit{S4, S5, S6, S8, S9, S13} and \textit{S15} the trained model performed better compared to the training set. Furthermore, the performance of the model on the test data for left out subjects \textit{S10} and \textit{S16} was similar to that on the training set. The discrepancies in the results of LOSO analysis can be attributed to many different reasons such as physical characteristics, emotional endurance, stress management skills, personality traits, and noise in the sensor data. %We analyzed the details of each participant to find out patterns that can explain the results of LOSO analysis, but we could not find any significant correlations. We did discover some possible reasons behind the individual performance for some subjects. 
For example, subject $S3$ was looking forward to stress conditions and was cheerful during data collection. Subject $S5$ might have fallen asleep during the first meditation phase and subject $S6$ had a stressful week, and the study was relaxing and not very stressful. Also, subject $S8$ already had a stressful day before the study and felt cold in the study room. These observations suggest that to better account for the differences between individuals towards the perception of stressful events and to build a general model for stress classification personalization of stress models is needed.

The authors in \cite{b3} also used LOSO for cross-validation and were able to achieve a classification accuracy of $95.21\%$ and f1-score of $0.94$ with a neural network trained on features data computed from all sensor modalities. In \cite{b7} the authors, computed various features from EDA data, and were able to achieve an average f1-score of $0.89$ with the XGBoost algorithm. Our LOSO analysis is based on just EDA data without any feature computing, and on average, across subjects, our method has the classification accuracy of $85.44\%$ and f1-score of $0.75$.

\vspace{-2.5mm}
\vspace{-2.5mm}
\vspace{-1.5mm}

To personalize the stress models for left-out subjects whose test set performance was lower than the training set, we re-trained the machine learning models on the left-out subject data. Starting from $1$ sample from the test set, we successively increased the number of samples used for re-training the model until the performance of the model on the test set was greater or equal to that on the original training set. Table \ref{tab:personalization_samples} shows the number of samples needed for each left-out subject and the final test set accuracy after re-training. The performance of the model on the test set increased significantly after re-training, suggesting we need personalized stress models for maximum performance.

\vspace{-3.5mm}
\begin{table}[htbp]
    \caption{Number of samples needed to personalize stress models for left out subjects with test accuracy lower than training accuracy.}
    \begin{center}
    % \resizebox{\linewidth}{!}{%
    \begin{tabular}{|c|c|c|c|c|}
    \hline
     Subject & \multicolumn{1}{|p{1.3cm}|}{\centering Original \\ Test Set \\ Accuracy} & \multicolumn{1}{|p{1.3cm}|}{\centering Total Samples} & \multicolumn{1}{|p{1.3cm}|}{\centering Re-training \\ Sample Size} & \multicolumn{1}{|p{1.3cm}|}{\centering Final \\ Test Set \\ Accuracy} \\
    \hline
    S2 & $76.8$ & $56$ & $43$ & $96.4$ \\
    \hline
    S3 & $67.9$ & $56$ & $56$ & $83.9$ \\
    \hline
    S7 & $84.5$ & $58$ & $40$ & $98.3$ \\
    \hline
    S11 & $66.1$ & $59$ & $52$ & $98.3$ \\
    \hline
    S14 & $55.9$ & $59$ & $42$ & $94.9$  \\
    \hline
    S17 & $57.4$ & $61$ & $43$ & $93.4$ \\
    \hline
    \end{tabular}
    % }
    \label{tab:personalization_samples}
    \end{center}
\end{table}
\vspace{-2.5mm}

% We can further shows that by adding few samples from each subject in the training set, we can improve the performance of the trained model on the test set. This will confirm we need personalization. 

% The outcomes of Michael's work that showed that stress is subjective can be linked here to present a strong case for personalized model for stress detection. 

% Bi-affective state classification (Stress vs Non-Stress) and Tri-affective state classification (Stress vs Amusement vs Baseline). Multi-affective state classification cases (Stress vs Amusement vs Baseline vs Meditation). The models achieved a classification accuracy of $95\%$ in the Bi-affective state, $85\%$ in Tri-affective state and $83\%$ in the Multi-affective state cases.

% For stress detection and classification we present results for bi-affective state (stress vs. normal) and tri-affective state (stress vs. normal vs. amusement) classifications. 

% \begin{enumerate}
%     \item Learning Rate
%     \item Filters, Kernel Size, Strides, Drop out percentage, Units, and number of layers of the CNN model
%     \item Size of window and percentage of the overlap
%     \item Number of epochs and batch size
%     \item percentage of train, test and validation set
% \end{enumerate}

\section{Conclusion}
In this work, we proposed a novel CNN-based stress detection and classification framework that uses raw EDA sensor data without feature computation and selection for affective states (stressed vs. normal vs. amusement) classification. We used the EDA data because EDA is found to be the best indicator of stress. Our approach can be adapted to include other sensor modalities for possible performance improvement also extended to other datasets. We also showed the need for a personalized stress model with our leave one subject analysis. Our approach is competitive with other state-of-the-art methods and does not suffer from many disadvantages such as feature computation and selection, multi-modal input data, and complex system design.

% \section*{References}

% \vspace{12pt}
% \color{red}
% IEEE conference templates contain guidance text for composing and formatting conference papers. Please ensure that all template text is removed from your conference paper prior to submission to the conference. Failure to remove the template text from your paper may result in your paper not being published.

\end{document}